\pdfoutput=1

\documentclass[11pt]{article}

\usepackage[]{acl}

\usepackage{times}
\usepackage{latexsym}

\usepackage[T1]{fontenc}

\usepackage[utf8]{inputenc}

\usepackage{microtype}

\usepackage{textcomp}

\usepackage{graphicx}
\graphicspath{ {./figs/} }

\usepackage{booktabs}

\usepackage{tabularx}

\usepackage{footmisc}

\usepackage{amsmath}

\usepackage{cleveref}

\crefformat{section}{\S#2#1#3} 
\crefformat{subsection}{\S#2#1#3}
\crefformat{subsubsection}{\S#2#1#3}

\DeclareMathOperator*{\argmax}{arg\,max}

% ...
\usepackage{booktabs, multirow} 
\usepackage{soul}               
\usepackage{array}

\usepackage{xspace}
\usepackage{soul}

\usepackage{cleveref}
\crefformat{section}{\S#2#1#3}
\crefformat{subsection}{\S#2#1#3}
\crefformat{subsubsection}{\S#2#1#3}
\crefrangeformat{section}{\S\S#3#1#4 to~#5#2#6}
\crefmultiformat{section}{\S\S#2#1#3}{ and~#2#1#3}{, #2#1#3}{ and~#2#1#3}
\usepackage{refstyle}
\usepackage{graphicx}
\Crefformat{figure}{#2Fig.~#1#3}
\Crefmultiformat{figure}{Figs.~#2#1#3}{ and~#2#1#3}{, #2#1#3}{ and~#2#1#3}
\Crefformat{table}{#2Tab.~#1#3}
\Crefmultiformat{table}{Tabs.~#2#1#3}{ and~#2#1#3}{, #2#1#3}{ and~#2#1#3}
\Crefformat{appendix}{Appx.~\S#2#1#3}
\crefformat{algorithm}{Alg.~#2#1#3}

\newcommand{\QT}[1]{``{#1}''\xspace}

\newcommand{\ie}{\textit{i}.\textit{e}.\ }

\newcommand{\asecref}[1]{Appendix \ref{#1}}
\newcommand{\tbref}[1]{Table \ref{#1}}

\newcommand{\dotieconcat}[2]{
  \text{\raisebox{.8ex}{$\smallfrown$}}%
}

\newcommand{\tsb}{\textsubscript}

\newcommand\papername{\textsc{ViPhy}\xspace}

\title{\papername: Probing \QT{Visible} Physical Commonsense Knowledge}

\author{ Shikhar Singh, Ehsan Qasemi, Muhao Chen
    \\
    University of Southern California
    \\
    Los Angeles, California, USA
    \\
    \texttt{\{ssingh43,qasemi,muhaoche\}@usc.edu}
    \\}

\begin{document}
\maketitle

\begin{abstract}



In recent years, vision-language models (VLMs) have shown remarkable performance on visual reasoning tasks (e.g. attributes, location).
While such tasks measure the requisite knowledge to ground and reason over a given visual instance, they do not, however, measure the ability of VLMs to retain and generalize such knowledge.
In this work, we evaluate their ability to acquire \QT{visible} physical knowledge -- the information that is easily accessible from images of static scenes, particularly across the dimensions of object color, size and space.
We build an automatic pipeline to derive a comprehensive knowledge resource for calibrating and probing these models.
Our results indicate a severe gap between model and human performance across all three tasks.
Furthermore, our caption pretrained baseline (CapBERT) significantly outperforms VLMs on both size and spatial tasks -- highlighting that despite sufficient access to ground language with visual modality, they struggle to retain such knowledge. 
The dataset and code are available at \url{https://github.com/Axe--/ViPhy}.





\end{abstract}







\section{Introduction}

The ability to reason and acquire knowledge from experience, while being intuitive for humans, has been a long-standing challenge for AI agents~\cite{mccarthy1960programs}.
Examples such as the color of grass, or the relative position of monitor and table, are formally regarded as commonsense knowledge~\cite{chi2005commonsense}.
The retention of such knowledge in humans is achievable due to the presence of long-term memory, broadly classified into \textit{episodic} and \textit{semantic} memory~\cite{tulving1972episodic, camina2017neuroanatomical}. 
While the former stores information pertaining to personal events, the latter is geared towards general, decontextualized knowledge.\footnote{For instance, memory of one's birthday cake is episodic, whereas knowing that most birthdays include a cake is part of semantic memory.}
Prior studies~\cite{greenberg2010interdependence} have acknowledged the interdependency between them, particularly the \textit{consolidation} of semantic knowledge from episodic memories -- aids humans acquire commonsense from experience. 

\begin{figure}[!t]
    \centering
    \includegraphics[width=0.38\textwidth]{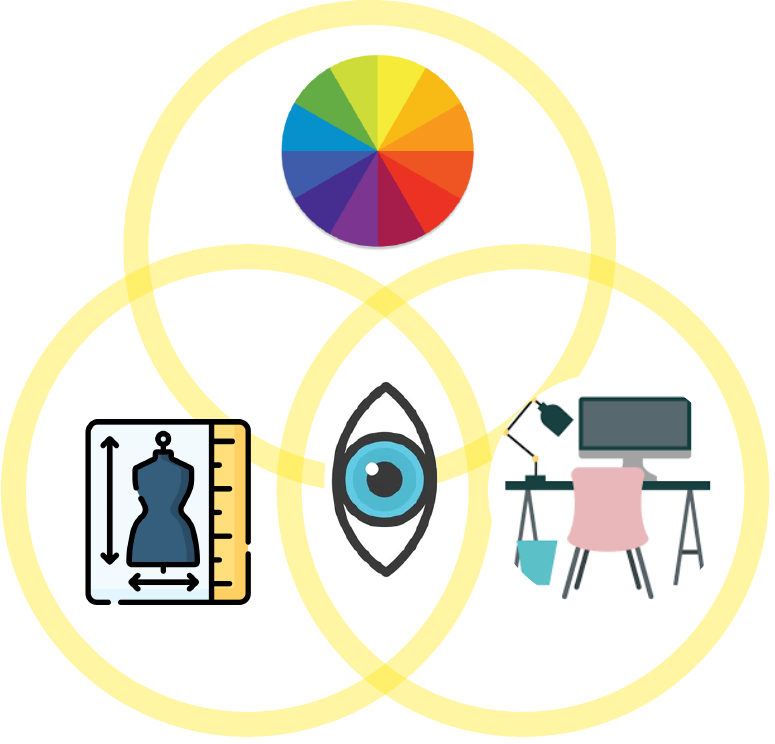}
    \caption{
        We propose \papername{} -- the trinity of visually accessible knowledge representing color, size and space.
    }
    \label{fig:intro}
\end{figure}

Pretrained language models~\cite{devlin2018bert, raffel2020exploring} have demonstrated the capacity to reason~\cite{wang2019superglue} and retain knowledge~\cite{petroni2019language, da2021analyzing}.
Likewise, vision-language models~\cite{lu2019vilbert, radford2021learning} driven by the availability of large-scale paired image-text datasets have shown strong performance on visual reasoning tasks~\cite{antol2015vqa, chen2015microsoft}.
While such tasks emphasize model's ability to draw inferences from a specific visual instance -- primarily to ground entities and reason about their attributes and relations, they do not, however, explicitly measure the consolidation of such knowledge.\footnote{
    An example of reasoning is counting bike wheels in an image, whereas knowing that bikes have two wheels is consolidation.}
In this work, we evaluate model's ability to recall aspects of grounding and reasoning tasks, regarded as commonsense knowledge.


Prior works have been largely directed towards probing language models pertaining to object properties such as weight, size, speed and affordance~\cite{forbes2017verb, forbes2019neural}.
Drawing upon the notion of world scopes~\cite{bisk2020experience}, we find that such datasets, albeit comprehensive across aspects of physical knowledge, are ideally suited for embodied agents capable of interacting with the physical environment.  
This motivates us to develop resources that better align with the world scope of existing AI systems, primarily vision-language models.

In this work, we introduce \papername, a \textbf{vi}sible \textbf{phy}sical commonsense dataset designed to probe aspects of physical knowledge that are easily accessible in images of static scenes.
Therefore, it can be argued that models pretrained on such data have sufficient access to the \QT{visible world}. 
We build a large-scale dataset along three dimensions of objects: (1) color, (2) size, and (3) space.
In contrast to prior works~\cite{paik2021world}, we bypass crowdsourced annotations in favor of an automated pipeline to derive a resource spanning 14k objects (30$\times$) from raw images. 
This is achieved by extracting object subtypes -- informed by visual context in images (e.g. kitchen sink).
We leverage image data, along with existing vision-language and depth perception models to develop \papername.

Beyond scale, we introduce a resource for probing spatial knowledge of common environments.
Albeit, one can reason along several types of spatial relations for a visual instance (e.g. a cat \textit{behind} a laptop; \citet{liu2022visual}) -- we find that mapping them to commonsense knowledge is non-trivial\footnote{
    Due to challenges in specifying the reference frame of the observer, canonical pose of the objects, and the situational nature of a scene.
}.  
We define spatial relations by selecting \QT{ground} as the observer, and specifying the relative elevation of objects under an allocentric reference frame~\cite{klatzky1998allocentric}.

We probe state-of-the-art models on \papername, and find significant gap across all three dimensions, compared to human performance.
Previous works~\cite{paik2021world, liu2022things} have corroborated the improvements from language grounding towards acquiring visual knowledge -- our results however, show a more nuanced picture.
While VLMs fair much better than LMs on recalling colors, our caption pretrained baseline (CapBERT) significantly outperforms VLMs on both size and spatial inference tasks. 
This highlights that despite access to visual modality, existing VLMs struggle to effectively consolidate such knowledge.

The contributions of this work can be as summarized as follows:
(1) We contribute with a comprehensive dataset, covering multiple aspects of visually accessible knowledge (\cref{sec:data}),
which is introduced through an automated pipeline to derive high-quality resource from images at scale (\cref{sec:pipeline});
(2) We conduct extensive benchmarking across several state-of-the-art language and vision-language models (\cref{sec:expt});
(3) We introduce a caption pretrained baseline that significantly outperforms state-of-the-art VLMs on several dimensions of \papername, highlighting of limitations of such models.





\begin{figure}[!t]
    \centering
    \includegraphics[width=0.4\textwidth]{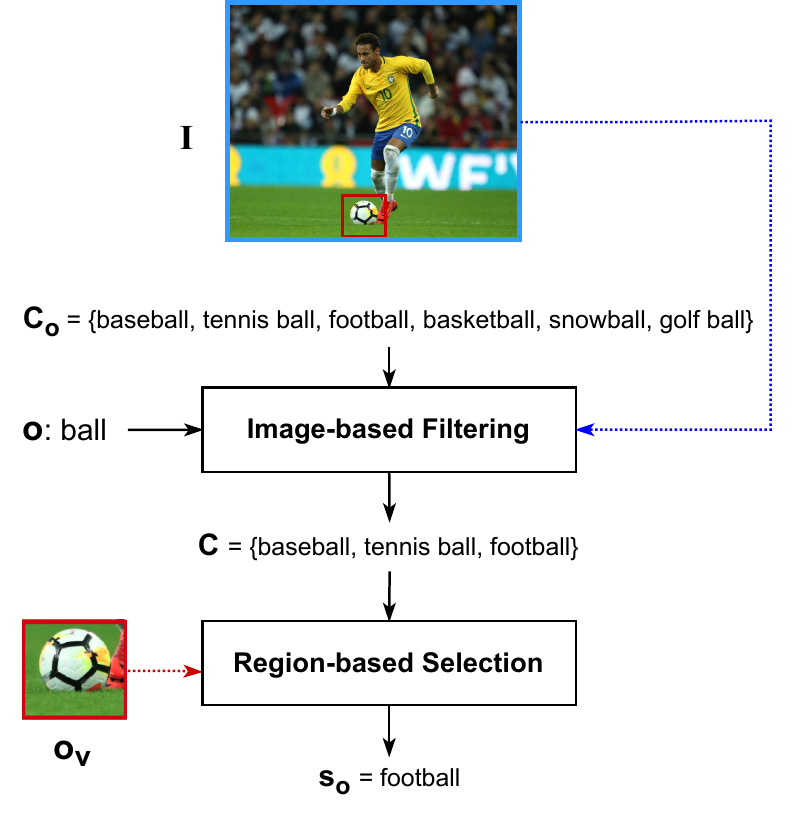}
    \caption{
        Subtype Selection Module: Given object $o$ in image $I$, assigns subtype $s_{o}$ from candidate set $C_{o}$.
    }
    ~\label{fig:subtype}
\vspace{-1em}
\end{figure}

\section{Pipeline}
\label{sec:pipeline}

\begin{figure*}[!t]
    \centering
    \includegraphics[width=0.95\textwidth]{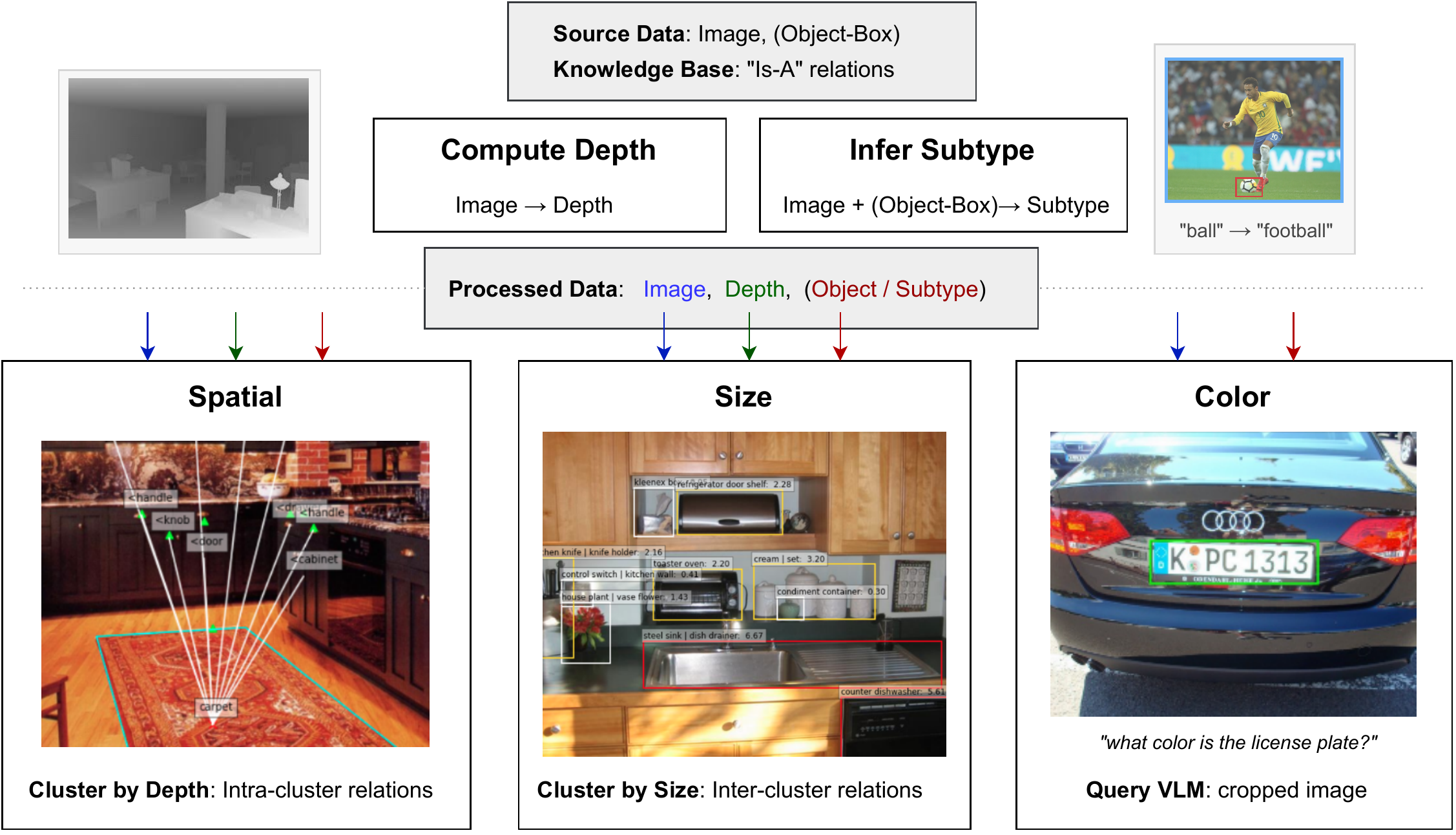}
    \caption{
        Pipeline Overview: The preprocessing stage computes the depth map for input image and re-annotates object regions with their subtype. It then independently extracts color, size and spatial knowledge.
    }
    ~\label{fig:pipeline}
\vspace{-1em}
\end{figure*}

We provide a conceptual overview of our pipeline for developing \papername, as illustrated in~\Cref{fig:pipeline}.
Given image and object regions as input\footnote{
    Available from either manual annotations or object detection / segmentation model (e.g. \citet{he2017mask}).
}, we substitute object names with their subtypes (\cref{subsec:pipe-subtype}), and compute the corresponding depth map.
The subtype candidates are acquired from image captions and knowledge bases.
After preprocessing, we independently extract color (\cref{subsec:pipe-color}), size (\cref{subsec:pipe-size}) and spatial knowledge (\cref{subsec:pipe-spatial}).


\subsection{Object Subtype}
\label{subsec:pipe-subtype}


While object recognition datasets consider a wide range of objects, such tasks do not necessitate fine-grained categories~\cite{zou2019object}.
However, object \textit{subtypes} inform attributes such as color, and help contextualize objects in absence of visual signals (e.g. office chair).
Albeit, subtypes are generally accessible from knowledge bases (KB), their coverage is often limited.\footnote{
    We report \char`\~60\% object name overlap between our collection and ConceptNet KB~\cite{speer2017conceptnet}.}
We extend this definition to include objects defined by visual context -- indicating event, location, state, part, etc. (Appendix~\Cref{atab:subtype-ex}).
For \textbf{subtype collection}, we parse the captions to build a set of object names.
We then employ suffix-based lexical matching to derive subtypes for each object, and merge with hyponyms from knowledge base. 
The resulting data represents a mapping between the object name and its candidate subtypes.

As our goal is to derive object attributes and relations directly from images, we design a \textbf{subtype selection} module to annotate the source image regions with the best subtype.
This is required since human annotators often abstract the object name to avoid redundancy when presented with visual context (example in Appendix~\Cref{afig:img-annot}), congruent with the maxim of quantity~\cite{grice1975logic}. 
Likewise, existing object detectors are not suited for open-vocabulary and fine-grained classification~\cite{minderer2022simple}.

The module is designed to query from subtype candidates using visual features. 
It employs a two-stage approach to filter candidates using image context, and select the best subtype with region-level features, as illustrated in~\Cref{fig:subtype}.
The visual and textual inputs are embedded using a dual stream vision-language model.
Formally, given the visual feature of the image $I$, textual features of the object $o$ and subtype candidates $C_{o}$, we extract the appropriate subtype as follows:

\vspace{-1.5em}
$$
    C = \{c | c \in C_{o}, sim(c, I) > sim(o, I)\} \cup o
$$

\noindent Here, $sim(.)$ is the cosine similarity.
Intuitively, since the object name is independent of visual context, it serves as an anchor for excluding subtypes that do not align with the contextual cues.
In the next stage, we incorporate visual features of the object region $o_{v}$, to query from filtered candidate set $C$, and compute the best subtype $s_{o}$:

\vspace{-1em}
$$
    s_{o} = \argmax_{c\in C} \hspace{0.5em} sim(o_{v}, c)
$$

\noindent The resulting dataset comprises of object-subtype mapping for bounding box regions in the image.

\subsection{Color}
\label{subsec:pipe-color}


Prior works~\cite{paik2021world} have relied on human annotations to acquire the color distribution of objects instead of inferring color from pixel values due to challenges such as lighting, shadow, segmentation, etc.
However, we argue that large-scale availability of images can mitigate potential noise associated with automated extraction.
Given the ubiquity of color attribute in visual reasoning tasks~\cite{antol2015vqa, hudson2019gqa}, we find that VLMs pretrained on such datasets are reliable for inferring color from images.
As object localization is decoupled from attribute recognition in the pipeline, the input to the VLM is simply the cropped image region, queried with a predefined textual prompt (detailed in \cref{para:impl-models}).

\subsection{Size}
\label{subsec:pipe-size}

To derive size relations, we consider co-occurring objects in a scene.
As objects in an image are expected to appear at varying depths, we approximate perceived size by including scene depth. 
Given an image, depth map and object-region annotations as inputs, the objects are clustered by size -- defined as the bounding box area scaled by mean depth of the region.
The sorted partitions are then used to derive inter-cluster relations. 
The object pair relations are aggregated across images.
The number of clusters are fixed for all instances.

\subsection{Spatial}
\label{subsec:pipe-spatial}

\begin{figure}[!t]
    \centering
    \includegraphics[width=0.45\textwidth]{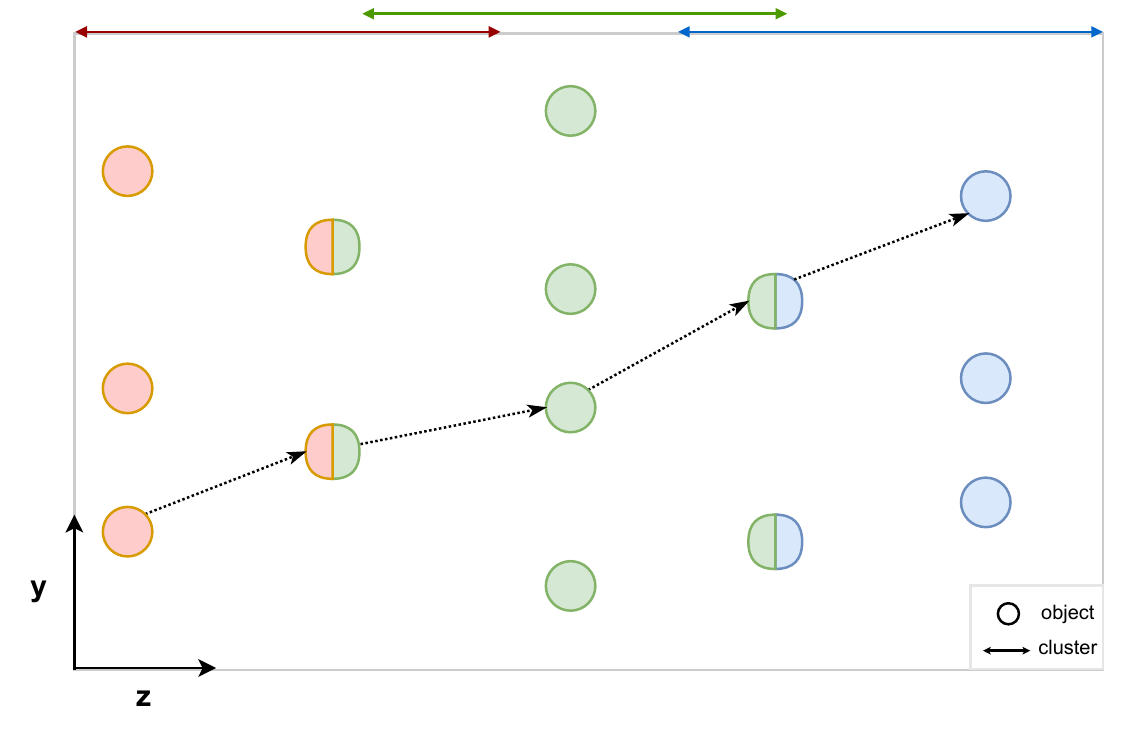}
    \caption{
        Illustrates transitive spatial relation, computed across partitions (ordered by depth). The y-axis denotes elevation, while the z-axis indicates depth. 
    }
    ~\label{fig:spatial}
\vspace{-1em}
\end{figure}

We define spatial knowledge as the relative elevation between objects, for a given scene type.
To infer these relations directly from image, however, is challenging as perspective projection of 3D world distorts the relative elevation due to variation in depth. 
We discount this distortion by partitioning the image by depth, and compute \textit{intra-cluster} object relations, \ie we discard the depth coordinate of objects that belong to the same cluster, and simply compare the relative elevation.
The \textit{inter-cluster} relations are derived transitively via overlapping partitions -- defined by objects with dual membership, as illustrated in~\Cref{fig:spatial}.
The spatial relations are aggregated across all images for a given scene type.
We detail the specifics of mapping object annotations to spatial relations in \asecref{asubsec:impl-spatial}.

\section{Dataset}
\label{sec:data}

This section details the specific data sources and models used to develop \papername, and dataset statistics (\cref{subsec:data-stats}).
We also describe the task format (\cref{subsec:data-task}) for each dimension.
Additional parameters related to dataset construction are provided in \asecref{asubsec:impl-params}.

\subsection{Construction}
\label{sec:impl}


\paragraph{Sources}
\label{para:impl-sources}

We leverage two datasets: (1) Visual Genome~\cite{krishna2017visual}, and (2) ADE20K~\cite{zhou2017scene}.
The dense captions in Visual Genome provide a broad coverage of object classes, making it a suitable resource for collecting subtype candidates.
For extracting hyponyms from knowledge base, we acquire "is-a" relations from ConceptNet~\cite{speer2017conceptnet}, and augment the subtype candidate set.
We extract spatial relations from ADE20K, as it provides images categorized by scene type -- primarily indoor environments with high object density: \textit{\{bedroom, bathroom, kitchen, living room, office\}}.


\paragraph{Models}
\label{para:impl-models}

To collect subtype candidates (as detailed in \cref{subsec:pipe-subtype}), we perform part-of-speech tagging to extract object names (noun) from caption data, using LSTM-CRF~\cite{akbik2018coling}.
For subtype selection, we employ UniCL~\cite{yang2022unified} -- designed for discriminative representations and broad semantic coverage of entities.
To compute depth map from monocular image, we use DPT~\cite{ranftl2021vision}.
To infer object color from image region (\cref{subsec:pipe-color}), we query OFA~\cite{wang2022ofa}, using the prompt template: “what color is the <object>?”.
The zero-shot predictions from OFA are mapped to the basic color set~\cite{berlin1991basic} as detailed in \asecref{asubsec:impl-pred2color}.


\subsection{Statistics}
\label{subsec:data-stats}

\tbref{tab:stats} summarizes statistics for \papername, comprising the number of objects, classes and instances for each dimension.
For multi-label tasks, we report the label cardinality, \ie number of labels for a sample.
We also indicate the number of objects with subtypes and subtype cardinality\footnote{Note: The subtype cardinality is only reported for objects with subtype.}. 
Note that while we extract size relations and color attributes from same source images (\cref{para:impl-sources}), we ignore \textit{contextual} subtypes for size relations, as they serve a limited role towards informing object size (e.g. rain coat).
However, as we only consider co-occurring objects, we implicitly incorporate context for objects in comparison, \ie help disambiguate word sense. 
We collect 7.1k smaller and 14.5k larger relations, and balance labels by including their complements.
The label distributions of color dataset is provided in Appendix \Cref{fig:viphy-coda-color}.


\begin{table}[!t]
    \centering \small
    \begin{tabular*}{0.42\textwidth}{c @{\extracolsep{\fill}} llc}
        \toprule
        \textbf{Subtype }   & objects               & 14k           \\
                            & objects with subtype  & 1.8k          \\ 
                            & subtype cardinality   & 7.73 (12.91)  \\  [0.1cm]
                            
        \textbf{Color}      & objects (instances)   & 14k \\
                            & classes               & 11 \\     
                            & label cardinality     & 2.42 (1.37)  \\ [0.1cm]
                            
        \textbf{Size}       & objects               & 1.6k \\
                            & instances             & 43k \\
                            & relations             & 2 \\      [0.1cm]
                            
        \textbf{Spatial}    & objects               & 300 \\
                            & instances             & 6.5k \\
                            & relations             & 3 \\
                            & label cardinality     & 1.28 (0.45)  \\ 
                            & scenes                & 5 \\
        \bottomrule
    \end{tabular*}
    \caption{Dataset statistics for \papername. 
            The cardinality is reported with mean and standard deviation.} \label{tab:stats}
\end{table}


\subsection{Task Setup}
\label{subsec:data-task}

The objective of \papername tasks is to measure the ability to recall physical knowledge pertaining to objects.
To probe models with textual prompts, we map the raw distribution of labels (acquired by our pipeline) to typical values, as detailed in \asecref{asubsec:impl-typical}.
In the \textbf{color} task, objects can have multiple labels from the set of 11 basic colors as defined in~\citet{berlin1991basic}: \textit{\{red, orange, yellow, brown, green, blue, purple, pink, white, gray, black\}}.
Likewise, we consider multiple labels for \textbf{spatial} relations from $\{\textit{below}, \textit{above}, \textit{similar level}\}$, conditioned on a scene type as mention in \cref{para:impl-sources}.
Lastly, \textbf{size} relations are mapped to a single label from $\{\textit{smaller}, \textit{larger}\}$. 


\section{Experiments}
\label{sec:expt}

We evaluate several state-of-the-art models under zero-shot and finetune settings (\cref{subsec:results}), and conduct further analysis of model performance (\cref{subsec:analysis}).
The datasets are partitioned into 20\% train, 10\% dev, 70\% test set.

\paragraph{Baselines} We consider the following language (LM) and vision-language (VLM) models:

\begin{itemize}
    \setlength\itemsep{0.2em}
    
    \item \textbf{LMs}: BERT~\cite{devlin2018bert}, RoBERTa~\cite{liu2019roberta}, DeBERTa~\cite{he2020deberta} and UnifiedQA~\cite{khashabi2020unifiedqa}.
    
    \item \textbf{VLMs}: VisualBERT~\cite{li2019visualbert}, ViLT~\cite{kim2021vilt}, CLIP~\cite{radford2021learning} and FLAVA~\cite{singh2022flava}.
\end{itemize}

\paragraph{CapBERT} In addition to the aforementioned baselines, we explore the following question:
To what degree does an LM pretrained only on image captions -- encodes visual knowledge.
In contrast to the standard corpora which comprises a broad range of knowledge domains, captions primarily describe the visual aspects of the world.
We build CapBERT by pretraining BERT\tsb{base} on captions from COCO~\cite{chen2015microsoft}, CC3M~\cite{sharma2018conceptual} and VG~\cite{krishna2017visual} datasets.

\paragraph{Finetuning} To evaluate under finetune setting, we train a linear classifier on top of the model's output, while rest of the weights are frozen.
We use Softmax Cross Entropy loss for single and multi-label setups, following \citet{mahajan2018exploring}.
All probes are finetuned for 50 epochs, with batch size of 8, using Adam optimizer~\cite{kingma2014adam} and a learning rate of $10^{-4}$.

\paragraph{Prompts}  
For probing LMs and VLMs, we provide manually designed textual prompt as input to the model.
The prompt templates for probing across color, size and spatial tasks, under zero-shot (ZS) and finetune (FT) settings are given in \tbref{tab:prompt}.
Besides models trained on the masked language objective\footnote{
    CLIP being the exception, cannot be evaluated under ZS setting. Under FT, it uses \texttt{EOS} instead of \texttt{CLS} token.}, 
the question-answering baseline (UnifiedQA) follows a common template\footnote{The prompt includes all classes as choices.} for both ZS and FT settings. 


\begin{table}[!tp]\centering
\small
\begin{tabularx}{0.5\textwidth}{lcX}\toprule
    \textbf{Task}   &\textbf{Setting}     &\textbf{Prompt} \\
    \midrule 
    Color           & ZS        & $O$ is of \texttt{[MASK]} color \\
                    & FT        & \texttt{[CLS]} color of $O$ \\
                    & QA        & What is the color of $O$? 
                                    (a) ..  (b) ..\\ [0.5cm]
    
    Size            & ZS        & $O_{1}$ is \texttt{[MASK]} than $O_{2}$ in size  \\
                    & FT        & \texttt{[CLS]} size of $O_{1}$ in comparison to $O_{2}$\\
                    & QA        & what is the size of $O_{1}$ in comparison to $O_{2}$? 
                                    (a) ..  (b) ..\\  [0.5cm]
    
    Spatial         & ZS        & in a $S$, the $O_{1}$ is located \texttt{[MASK]} the $O_{2}$ \\
                    & FT        & \texttt{[CLS]} in a $S$, the $O_{1}$ is located in comparison to $O_{2}$  \\
                    & QA        & in a $S$, where is $O_{1}$ is located in comparison to $O_{2}$? 
                                (a) .. (b) .. \\
    \bottomrule
\end{tabularx}
\caption{Prompt templates across tasks and evaluation settings. Here, \textit{O}, \textit{R} and \textit{S} are placeholders for object, relation and scene type respectively.}\label{tab:prompt}
\end{table}

\paragraph{Metrics} We introduce the following metrics for measuring task performance under multi-label setting:

\begin{itemize}
    \item Relaxed Accuracy (R-Acc) -- The prediction ($P$) is accurate if the most probable label belongs to the ground-truth labels ($T$).
    $$
        RA = \sum_{i \in D} \frac{[l_i \cap T_i] \land [l_i = \argmax_{l} P_{i}(l)]}{|D|}
    $$
    \item True Confidence (Conf) -- The sum of predicted probabilities for labels in the ground-truth set.
    $$
        C = \sum_{i \in D} \frac{\sum_{l \in T_i} P_{i}(l)}{|D|}
    $$
\end{itemize}

\vspace{-0.5em}
\noindent Here, $D$ denotes samples in the evaluation set. In addition to the aforementioned metrics, we also report the macro-averaged F1-score (F1).


\paragraph{Human Performance} To provide an upper bound on \papername tasks, we use  CoDa~\cite{paik2021world} for color -- computed over 432 overlapping objects\footnote{The label distributions of \papername and CoDa are provided in Appendix -- \Cref{fig:viphy-coda-color} and \Cref{fig:viphy-coda-cardinality}}.
For size and spatial dataset, we internally evaluate 100 relations with two annotators (authors) and report the average score.

\subsection{Results}
\label{subsec:results}

\paragraph{Zero-Shot} We report zero-shot performance using R-Acc metric, across all tasks in \tbref{tab:zero-shot-results}.
For spatial task, we only consider two labels from $\{above, below\}$, due to the limitation of single word masking in selected baselines.
We observe significant variations in model performance across tasks, with VLMs (VisualBERT) performing worse than their LM counterparts -- underscoring the challenges of manual prompting~\cite{jiang2020can}.
The best scoring baseline (UnifiedQA) falls at least 30\% points below human scores. 


\begin{table}[h]\centering
    \small
        \begin{tabularx}{0.4\textwidth}{l @{\extracolsep{\fill}} ccc}
        \toprule
        \textbf{Model}          &\textbf{Color} &\textbf{Size} & \textbf{Spatial} \\
            \midrule
            BERT\tsb{large}         & 48.39 & 44.61 & 20.96 \\
            RoBERTa\tsb{large}      & 0.59  & 47.01 & 17.52 \\
            UnifiedQA\tsb{large}    & 51.00 & 51.76 & 63.04 \\
            VisualBERT              & 9.06  & 24.91 & 9.57 \\
            \cmidrule{1-4}
            Human                   & 97.45 & 94.32 & 95.12 \\
        \bottomrule
    \end{tabularx}
    \caption{Zero-shot results (R-Acc) across all tasks.}
    \label{tab:zero-shot-results}
\end{table}

\paragraph{Finetune} When compared to zero-shot results, we report improved  calibration under finetuned probing, as evident from results on color (\Cref{tab:color-results}), size (\Cref{tab:size-results}) and spatial tasks (\Cref{tab:spatial-results}).
We find that VLMs score higher than LMs -- specifically their \QT{caption-only} counterpart (CapBERT) on the color task. 
These results hint at the role of color attribute in grounding entities. 
However, CapBERT outperforms VLMs on both size and spatial tasks, implying that despite having access to visual representations, VLMs do not retain such relational knowledge as effectively. 
Lastly, CapBERT outperforming other LMs is likely due to the domain similarity between the pretraining source and the evaluation tasks\footnote{For instance, a prepositional phrase can convey both abstract (\textit{on schedule}) and physical (\textit{on table}) relations, with captions predominantly containing the latter.}.

\begin{table}[!t]
    \centering
    \small
    \begin{tabularx}{0.4\textwidth}{l @{\extracolsep{\fill}} ccc}
        \toprule
        \textbf{Model} &\textbf{R-Acc} &\textbf{Conf} & \textbf{F1} \\
            \midrule
            CapBERT             & 70.45 & 58.55 & 40.91  \\
            BERT\tsb{base}      & 66.87 & 55.59 & 30.94  \\
            RoBERTa\tsb{large}  & 55.95 & 49.28 & 20.93  \\
            UnifiedQA\tsb{large}& 62.34 & -     & - \\
            DeBERTa\tsb{xxl}    & 72.74 & 59.59 & 36.33  \\
            \cmidrule[0.1pt]{1-4}
            VisualBERT          & 66.22 & 50.99 & 24.46  \\
            ViLT                & 64.83 & 53.92 & 30.27  \\
            FLAVA               & 76.33 & 62.84 & 38.74  \\
            CLIP                & \textbf{79.96} & 65.50 & 49.54  \\
            \cmidrule{1-4}
            Human\tsb{ CoDa}    & 97.45 & 78.65 & 72.12  \\
        \bottomrule
    \end{tabularx}
    \caption{Color results.}
    \label{tab:color-results}
\end{table}

\begin{table}[!t]\centering
    \small
    \begin{tabularx}{0.4\textwidth}{l @{\extracolsep{\fill}} ccc}
        \toprule
        \textbf{Model} &\textbf{R-Acc} &\textbf{Conf} & \textbf{F1} \\
            \midrule
            CapBERT             & \textbf{69.93}    & 62.09 & 60.78  \\
            BERT\tsb{base}      & 67.25 & 59.91 & 61.34  \\
            RoBERTa\tsb{large}  & 54.88 & 58.40 & 58.88  \\
            UnifiedQA\tsb{large}& 62.04 & -     & -     \\
            DeBERTa\tsb{xxl}    & 62.30 & 61.27 & 60.54  \\
            \cmidrule[0.1pt]{1-4}
            VisualBERT          & 63.08 & 58.40 & 58.88  \\
            ViLT                & 65.78 & 60.28 & 59.80  \\
            FLAVA               & 63.71 & 61.06 & 60.56 \\
            CLIP                & 65.10 & 63.56 & 62.26  \\
            \cmidrule{1-4}
            Human               & 95.12 & -     & 87.42 \\
        \bottomrule
    \end{tabularx}
    \caption{Spatial results.}\label{tab:spatial-results}
\end{table}

\begin{table}[!t]\centering
    \small
    \begin{tabularx}{0.45\textwidth}{l @{\extracolsep{\fill}} ccc}
        \toprule
        \textbf{Model} &\textbf{Standard} &\textbf{Subtype} & \textbf{Transitive} \\
            \midrule
            CapBERT             & \textbf{83.69}  & \textbf{79.14} & \textbf{91.82} \\
            \cmidrule[0.1pt]{1-4}
            BERT\tsb{base}      & 78.35 & 72.28 & 77.29 \\
            RoBERTa\tsb{large}  & 65.23 & 57.12 & 69.31 \\
            UnifiedQA\tsb{large}& 62.20 & 60.66 & 90.78 \\
            DeBERTa\tsb{xxl}    & 74.73 & 66.88 & 69.79 \\
            \cmidrule[0.1pt]{1-4}
            \textbf{LM}\tsb{average}     & 69.37 & 64.23 & 74.54 \\
            \cmidrule[0.5pt]{1-4}
            VisualBERT          & 76.99 & 64.00 & 77.69 \\
            ViLT                & 78.54 & 57.32 & 86.18 \\
            FLAVA               & 82.67 & 69.54 & 81.78 \\
            CLIP                & 75.43 & 66.56 & 72.48 \\
            \cmidrule[0.1pt]{1-4}
            \textbf{VLM}\tsb{average}    & 79.15 & 64.35 & 79.53 \\
            \cmidrule[0.9pt]{1-4}
            Human               & 94.32 & - &- \\
        \bottomrule
    \end{tabularx}
    \caption{Size results reported across different evaluation sets, measured by accuracy (random baseline: 50\%).}\label{tab:size-results}.
\end{table}

\subsection{Analysis}
\label{subsec:analysis}

\begin{figure}[!t]
    \centering
    \includegraphics[width=0.45\textwidth]{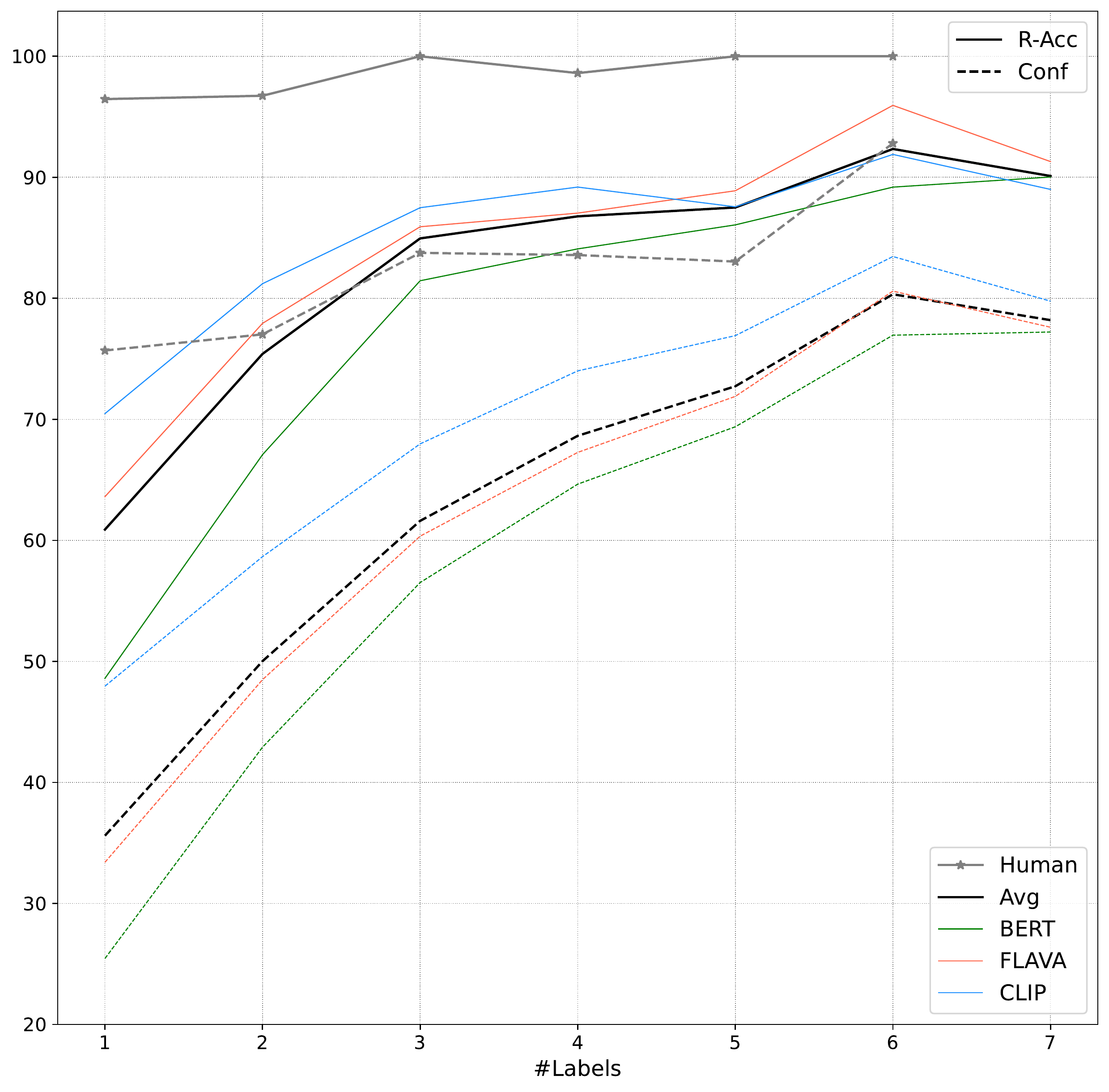}
    \caption{
        Effect of label cardinality (x-axis) on color prediction, 
        as measured by R-Acc and Conf. The \textit{Avg} curves (black) indicate average model performance.
    }
    \label{fig:color-cardinality-result}
\vspace{-1em}
\end{figure}

\paragraph{Color: Cardinality} We further analyze model performance with respect to label cardinality (i.e. number of ground-truth colors for an object), by grouping objects accordingly.
As shown in \Cref{fig:color-cardinality-result}, we report results for three baselines, their average, along with human scores. 
While the performance is expected to increase with the cardinality\footnote{R-Acc \& Conf are 1, when cardinality is 11.}, we notice an inconsistency between model and human scores.
In particular, while difference in overall confidence scores (as inferred from \tbref{tab:color-results}) for human and the model average is \char`\~18\%, the relative differences between the two -- ranges from \char`\~12\% (x = 6) to \char`\~40\% (x = 1), where x-axis denotes the label cardinality.
While color influences object perception in humans~\cite{gegenfurtner2000sensory}, these results show that VLMs do not ascribe a similar degree of saliency to color, especially for uni-color objects (\ie cardinality of one).


\paragraph{Size: Transitivity}

As the size dataset\footnote{
    For reference, the \#instances for size evaluation sets are as follows -- standard: 30k, subtype: 23k, transitive: 20k.} 
is composed of frequently co-occurring object pairs, we intend to evaluate models ability to infer relative size for objects linked transitively across scenes. 
We build a new evaluation set comprising transitive relations from the standard size dataset, ensuring no overlapping instances between the two.
The results (\Cref{tab:size-results}) indicate that LMs (on average) improve by significant margin, compared to VLMs.
While the improvements on the evaluation set can be partially attributed to objects on the relatively extreme ends of size clusters being paired up, they are able to generalize on transitive relations.

\paragraph{Size: Subtypes}

While contextual qualifiers tend to inform the typicality of color attribute, their effect on size is likely inconsequential.
Therefore, models should retain their performance with reference to the standard set.
We test this hypothesis by creating an evaluation set comprising contextual subtypes for objects in the standard test set.
While the addition of subtypes leads to performance drop across all models (\Cref{tab:size-results}), we find that LMs are more robust in comparison to VLMs.







\section{Related Works}

\paragraph{Physical Commonsense}

Recent years have witnessed a renewed interest in studying commonsense, primarily via natural language benchmarks~\cite{talmor2019commonsenseqa, singh2021com2sense}.
Specific works have evaluated the language models on their ability to reason about physical commonsense~\cite{bisk2020piqa, paco}.
Others have identified reporting bias as a potential bottleneck for models pretrained on text corpora towards acquiring physical knowledge~\cite{forbes2019neural, paik2021world}.
In this work, we direct our focus towards vision-language models pretrained on large paired image-text datasets, and evaluate them on visually accessible commonsense knowledge. 
While prior works have probed knowledge pertaining to color~\cite{paik2021world} and size~\cite{talmor2020olmpics}, their coverage of objects is severely limited in comparison to \papername (30$\times$).

Recently, \citet{liu2022visual} have comprehensively evaluated spatial reasoning for 65 relation types, given an image and a caption as inputs.
In contrast, \papername measures the ability to recall spatial commonsense. 
Additionally, whereas \citet{liu2022things} have probed spatial knowledge for human-object interaction (224 instances) under 15 action types (e.g. driving, cooking), we consider the spatial layout of objects across scene types over 6k instances, independent of events.

\paragraph{Vision-Language Resources}

While image classification~\cite{deng2009imagenet} can be construed as one of the earliest attempts at bridging vision and language, recent years have witnessed a plethora of resources.
Visual reasoning tasks have been directed towards object attributes~\cite{antol2015vqa}, activities~\cite{chen2015microsoft}, as well as social~\cite{zellers2019recognition} and temporal commonsense~\cite{fu-etal-2022-theres}. 
Recently, VLMs~\cite{lu2019vilbert, li2020oscar, radford2021learning} have demonstrated strong performance on such tasks.
These works evaluate the requisite knowledge to reason about a specific instance, \papername in contrast probes the knowledge retained in the absence of visual context, \ie generalized from instances.

\paragraph{Knowledge in LMs}

Recent advancements in language models~\cite{devlin2018bert, raffel2020exploring}, pretrained on large corpora, has led to significant improvements across several reasoning tasks~\cite{wang2019superglue}.
Prior works have also highlighted the capacity of these models to acquire several types of knowledge such as factual~\cite{petroni2019language, roberts2020much}, instructional~\cite{huang2022language} and commonsense~\cite{da2021analyzing}.
In this work, we study to what degree do their vision-language analogs (VLMs) -- driven by the availability of massive paired image-text datasets, retain information that is easily accessible in images.

\section{Conclusion}

We present \papername, a large scale resource for probing \QT{visible} physical knowledge -- information easily accessible from images of static scenes, across dimensions of color, size and space.
We design an automated pipeline to extract and consolidate such knowledge facts from images, and introduce a new resource for evaluating spatial knowledge of common environments.
Our benchmarking evaluation highlights a huge gap between model and human performance across all three tasks.
Furthermore, while prior works have reported VLMs to be more effective, our caption pretrained baseline (CapBERT) significantly outperforms VLMs on the ability to recall size and spatial knowledge.
These results underscore that despite access to visual modality, existing VLMs struggle to retain visual knowledge as effectively.




\section*{Ethical Implications}

We build \papername from existing images from crowd-verified visual datasets which have been identified to lack geographical diversity, often limited to scenes from Europe and North America~\cite{shankar2017no}. 
Furthermore, such datasets are subjected to several kinds of biases at different stages of collection and annotation such as selection bias, framing bias and observer bias~\cite{fabbrizzi2022survey}.
Therefore, its likely that such biases will be reflected in our dataset as well.
As we also report benchmarking results on \papername, the model performance may not be reflected as accurately on knowledge pertaining to different geographical and cultural backgrounds, as studied in \citet{yin2021broaden}. 
Lastly, our proposed resource is limited to English, and thus excludes any considerations for multilingual models~\cite{yin2022geomlama}.

\section*{Acknowledgement}

This work is supported in part by the DARPA MCS program under Contract No. N660011924033 with the United States Office Of Naval Research.

\bibliography{references, related}
\bibliographystyle{acl_natbib}

\newpage
\appendix

\section*{Appendix}

\section{Implementation Details}
\label{asec:impl}

\subsection{Pipeline Parameters}
\label{asubsec:impl-params}

To cluster objects for computing relative size, we use Jenks Natural Breaks~\cite{jenks1967data}, with \#clusters = 5 following the manual groupings of object sizes in \citet{liu2022things}.
We also experimented with \#clusters = 3, but qualitatively observed less optimal clusters. 
In spatial module, we create 3 partitions of uniform size and overlap. 

\subsection{Typical Labels}
\label{asubsec:impl-typical}

We derive typical labels from the raw label probabilities ($C$) by filtering classes as per a predefined threshold $p_{min}$, that can be interpreted as either noise or rare occurrence.
Formally, we apply the filter as follows: $C = \{(c, p) | p > p_{min}, (c, p) \in C\}$.
Here, $p_{min}$ is defined as:
\[ \begin{cases} 
      10\% & 4 \leq |C| \leq 11 \\
      20\% & |C| = 3 \\
      30\% & |C| = 2 \\ 
   \end{cases}
\]

The resulting distribution is re-normalized and the filtering step is applied recursively.

\subsection{Defining Spatial Relations}
\label{asubsec:impl-spatial}

\begin{figure}[ht]
    \centering
    \includegraphics[width=0.4\textwidth]{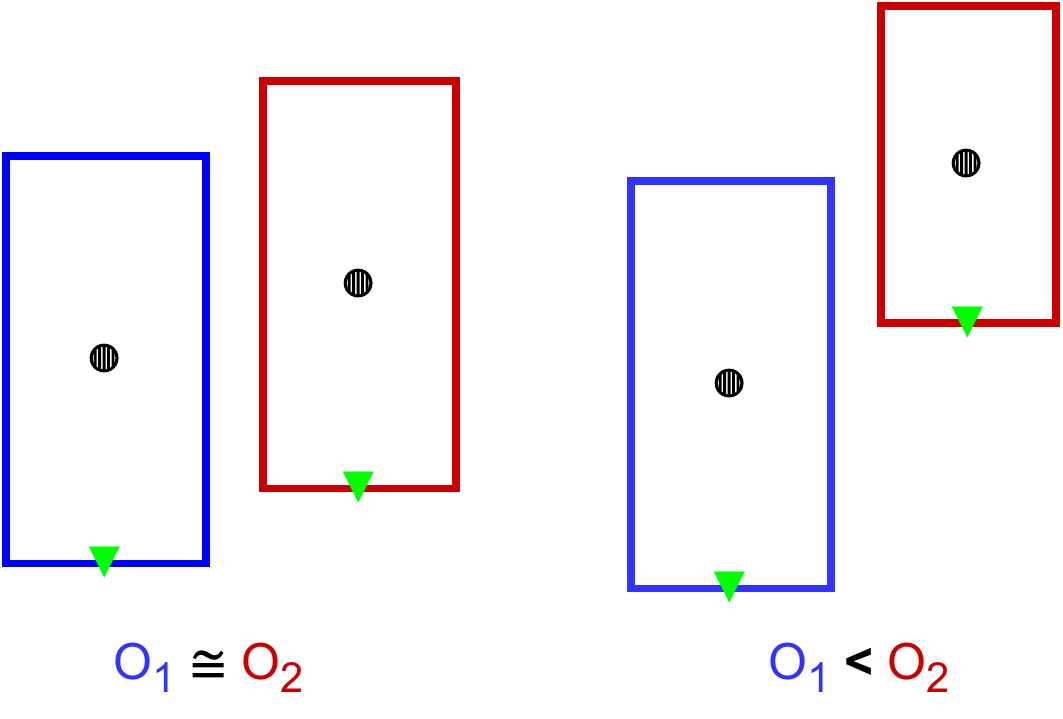}
    \caption{
        Illustrates our definition of spatial relation from raw annotations.
    }
    \label{afig:spatial-rel}
\vspace{-1em}
\end{figure}

Our objective is to map the raw coordinates in an image for two objects to discrete relations. 
We define a simple set of rules to convey \textit{above} and \textit{similar} level, from object annotations.
Given bounding box or polygon mask, we first compute its centroid along with the lowest point.
We then compare the y-coordinates of objects as illustrated in \Cref{afig:spatial-rel}.
If an object's lowest point is above the other's centroid, we map it to \textit{above}, else \textit{similar} level.

\subsection{Prediction to Basic Color}
\label{asubsec:impl-pred2color}

We use OFA in our pipeline, and map generated text to the basic color set as shown in \tbref{atab:pred2color}.
If the model predicts multiple colors, we assign each of them to the object instance.

\begin{table}[ht]
    \centering
    \small
    \begin{tabular}{ll}
        \toprule
        \textbf{Basic Color}    &   \textbf{Raw Predicted Terms} \\ 
        \midrule        
            Yellow  & \textit{gold, golden, blonde, beige, peach, cream}  \\ [0.1cm]
            Brown   & \textit{wooden, tan, beige, bronze, copper} \\ [0.1cm]
            Gray    & \textit{grey, silver, metal, steel} \\ [0.1cm]
            Pink    & \textit{peach} \\ [0.1cm]
            Purple  & \textit{violet} \\ [0.1cm]
            Red     & \textit{maroon} \\ [0.1cm]
            Green   & \textit{teal} \\ [0.1cm]
            Blue    & \textit{teal, turquoise} \\ [0.1cm]
        \bottomrule
    \end{tabular}
    \caption{Mapping between raw predictions (OFA) and basic color terms.}
    \label{atab:pred2color}
\end{table}



\begin{table}[ht]
    \centering
    \small
    \begin{tabular}{p{0.1\textwidth}l}
        \toprule
        \textbf{Context}    &   \textbf{Examples} \\ 
        \midrule        
            Event           &   \textit{wedding cake, bathing soap}  \\ [0.1cm]
            Location        &   \textit{kitchen sink, street lamp} \\ [0.1cm]
            State           &   \textit{bare tree, sliced apple} \\ [0.1cm]
            Part            &   \textit{piano key, bike wheel} \\ [0.1cm]
        \bottomrule
    \end{tabular}
    \caption{Context-based subtype examples}\label{atab:subtype-ex}
\end{table}

\begin{figure}[ht]
    \centering
    \includegraphics[width=0.5\textwidth]{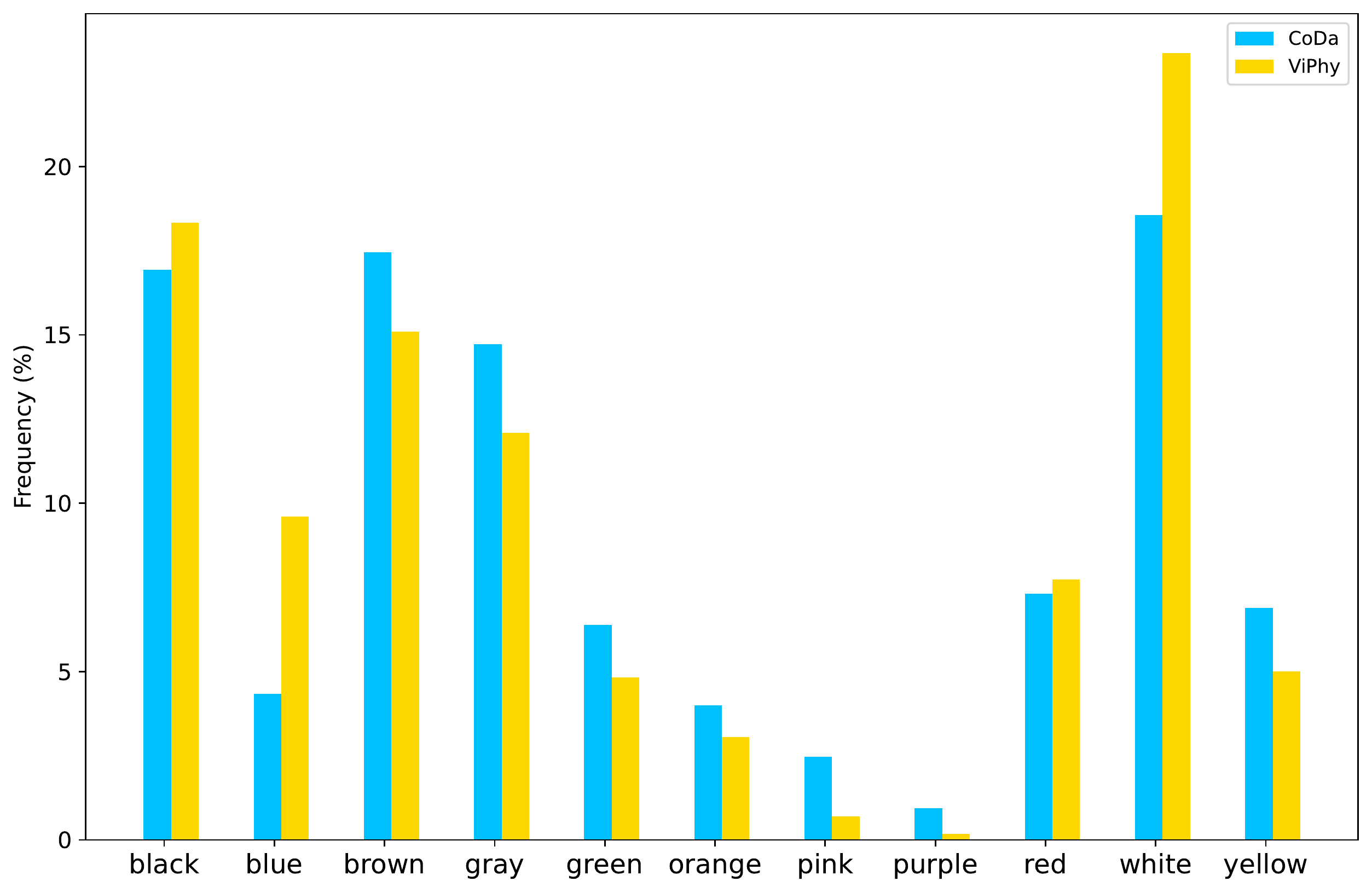}
    \caption{
        Color distribution of objects in \papername and CoDa. 
    }
    \label{fig:viphy-coda-color}
\vspace{-1em}
\end{figure}

\begin{figure}[ht]
    \centering
    \includegraphics[width=0.5\textwidth]{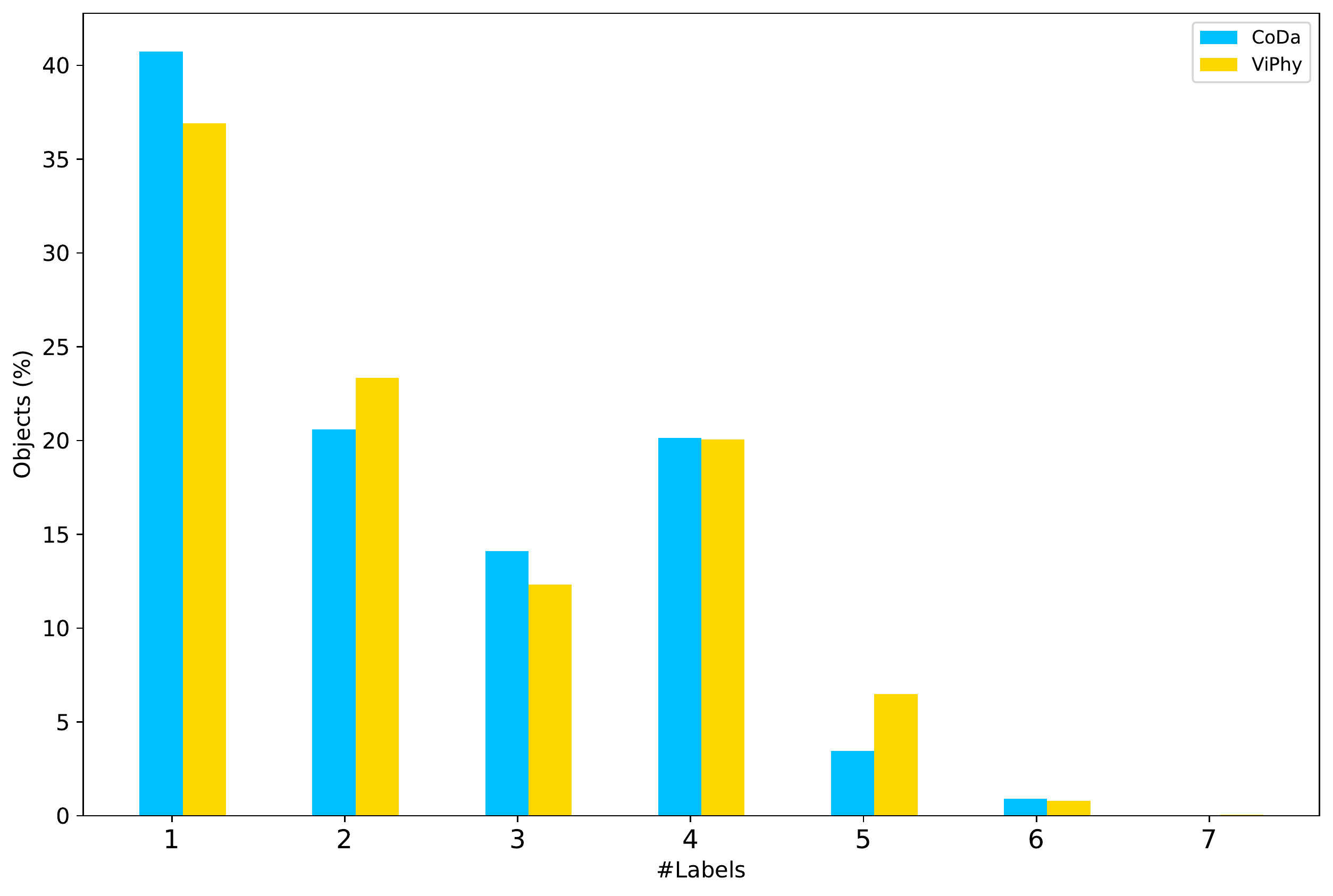}
    \caption{
        Color distribution with respect to label cardinality in \papername and CoDa.
    }
    \label{fig:viphy-coda-cardinality}
\vspace{-1em}
\end{figure}

\begin{figure}[ht]
    \centering
    \includegraphics[width=0.45\textwidth]{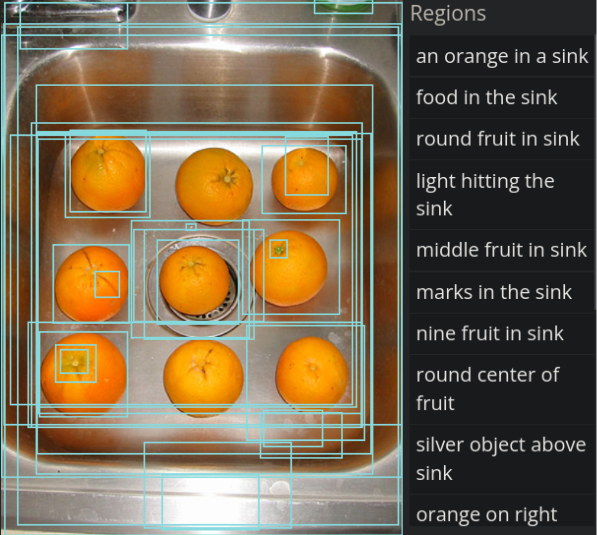}
    \caption{
        A sample image and corresponding captions from Visual Genome dataset.
        Illustrates how humans omit the subtype \textit{kitchen} sink, when annotating images.
    }
    \label{afig:img-annot}
\vspace{-1em}
\end{figure}

\begin{figure}[ht]
    \centering
    \includegraphics[width=0.5\textwidth]{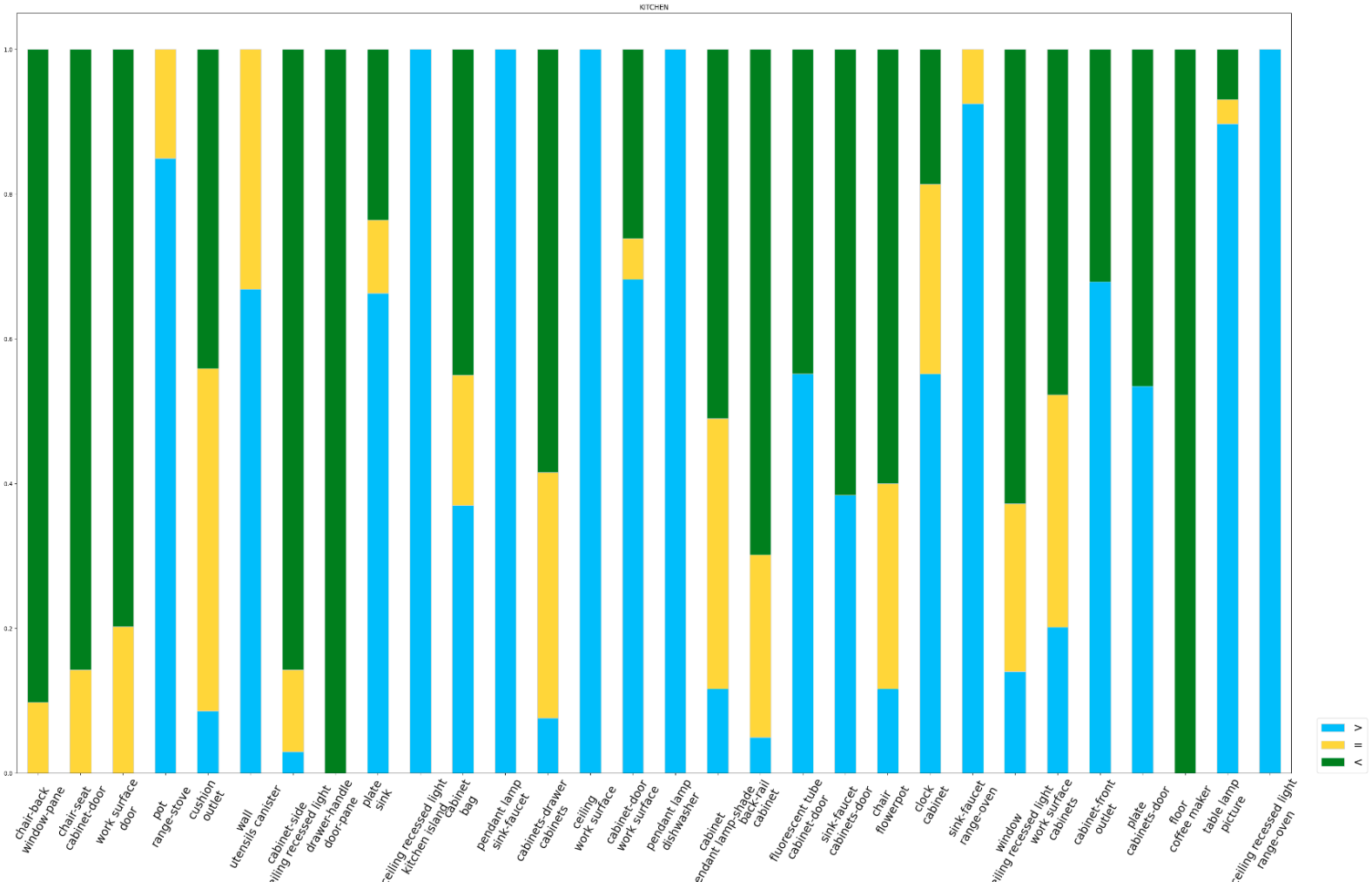}
    \caption{
        Spatial relation distribution for object pairs in \textit{kitchen} scene from \papername.
    }
    \label{fig:color-dist}
\vspace{-1em}
\end{figure}

\begin{figure*}[!t]
    \centering
    \includegraphics[width=0.95\textwidth]{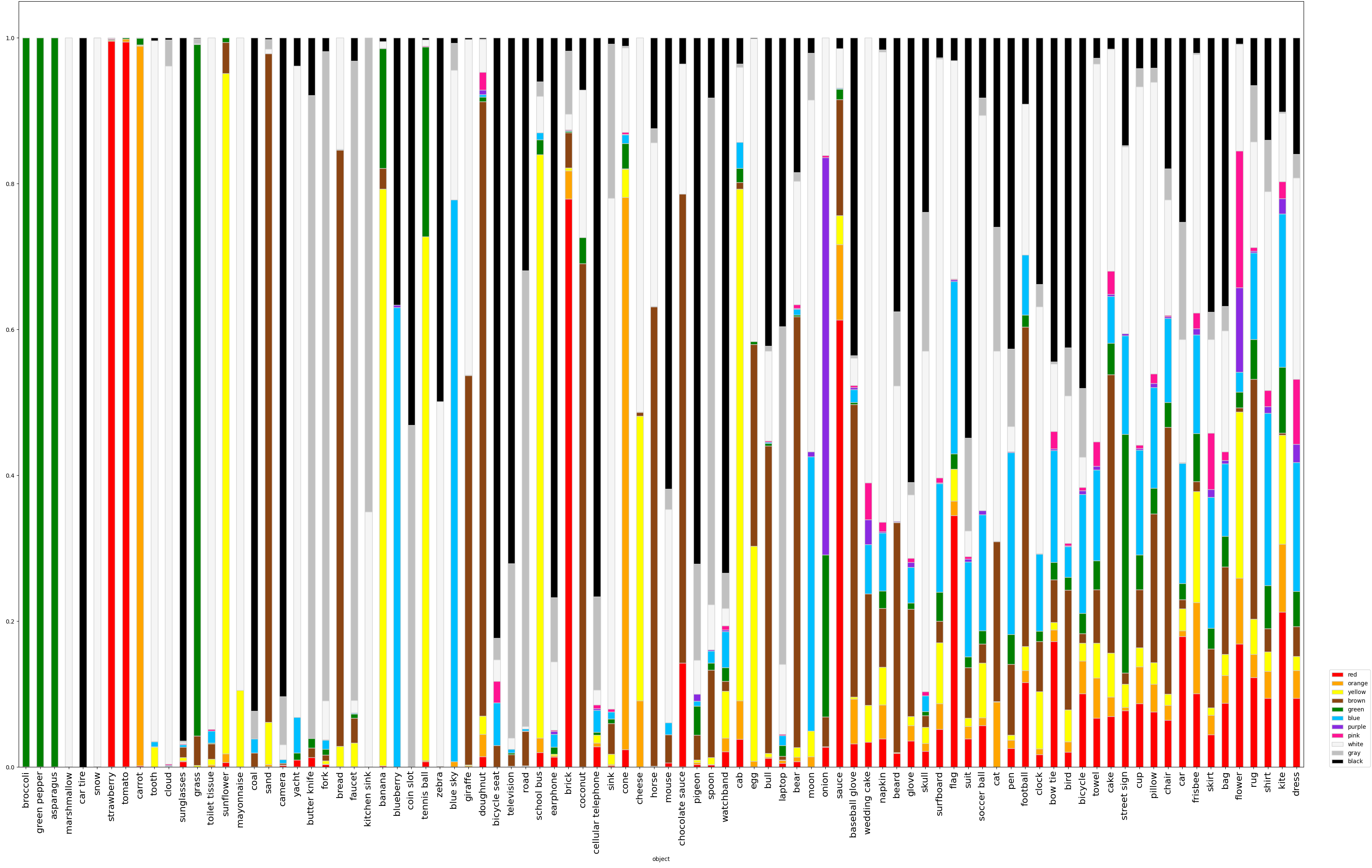}
    \caption{
        Color distribution for 90 objects from \papername, sorted by entropy.
    }
    \label{fig:color-dist}
\vspace{-1em}
\end{figure*}


\end{document}